\documentclass[10pt,twocolumn,letterpaper]{article}

\usepackage{cvpr}              

\usepackage{graphicx}
\usepackage{amsmath}
\usepackage{amssymb}
\usepackage{booktabs}

\usepackage{xcolor}
\usepackage{amsmath}
\usepackage{pifont}
\usepackage{graphicx}

\usepackage{overpic}

\usepackage{amssymb}
\usepackage{booktabs}

\usepackage{multirow}
\newcommand{\tabincell}[2]{\begin{tabular}{@{}#1@{}}#2\end{tabular}}
\usepackage{makecell}

\usepackage{balance}
\usepackage{bm}

\usepackage{colortbl}
\definecolor{mygreen}{RGB}{239,255,246}

\usepackage{balance}

\usepackage[pagebackref,breaklinks,colorlinks]{hyperref}

\usepackage[capitalize]{cleveref}
\crefname{section}{Sec.}{Secs.}
\Crefname{section}{Section}{Sections}
\Crefname{table}{Table}{Tables}
\crefname{table}{Tab.}{Tabs.}

\def\cvprPaperID{11424} 
\def\confName{CVPR}
\def\confYear{2023}

\begin{document}

\title{Adaptive Data-Free Quantization}

\author{Biao Qian, Yang Wang\thanks{Yang Wang is the corresponding author.} , Richang Hong, Meng Wang\\
Key Laboratory of Knowledge Engineering with Big Data, Ministry of Education,\\
School of Computer Science and Information Engineering,\\
Hefei University of Technology, China\\
{\tt\small yangwang@hfut.edu.cn, \{hfutqian,hongrc.hfut,eric.mengwang\}@gmail.com}
}
\maketitle

\begin{abstract}

Data-free quantization (DFQ) recovers the performance of quantized network (Q) without the original data, but generates the fake sample via a generator (G) by learning from full-precision network (P), which, however, is totally independent of Q, overlooking the adaptability of the knowledge from generated samples, i.e., informative or not to the learning process of Q, resulting into the overflow of generalization error. Building on this, several critical questions — how to measure the sample adaptability to Q under varied bit-width scenarios? whether the largest adaptability is the best? how to generate the samples with adaptive adaptability to improve Q’s generalization?  To answer the above questions, in this paper, we propose an \underline{Ada}ptive \underline{D}ata-\underline{F}ree \underline{Q}uantization (AdaDFQ) method, which revisits DFQ from a zero-sum game perspective upon the sample adaptability between two players — a generator and a quantized network. Following this viewpoint, we further define the disagreement and agreement samples to form two boundaries, where the margin is optimized to adaptively regulate the adaptability of generated samples to Q, so as to address the over-and-under fitting issues. Our AdaDFQ reveals : 1) the largest adaptability is NOT the best for sample generation to benefit Q's generalization; 2) the knowledge of the generated sample should not be informative to Q only, but also related to the category and distribution information of the training data for P. The theoretical and empirical analysis validate the advantages of AdaDFQ over the state-of-the-arts. Our code is available at https://github.com/hfutqian/AdaDFQ.
\vspace{-0.8em}
\end{abstract}


\section{Introduction}
\label{intro}
Deep Neural Networks (DNNs) have encountered great challenges when applied to the resource-constrained devices, owing to the increasing demands for computing and storage resources. Network quantization \cite{lin2016fixed,jacob2018quantization}, a promising approach to improve the efficiency of DNNs, reduces the model size by mapping the floating-point weights and activations to low-bit ones.  Quantization methods generally recover the performance loss from the quantization errors, such as fine-tuning or calibration operations with the original training data.\footnotetext{$^1$3-bit and 5-bit precision are representative for low-bit and high-bit cases, respectively, particularly: 3-bit quantization actually leads to a huge performance loss, which is a major challenge for the existing DFQ methods; while 5-bit or higher-bit quantization usually causes a small performance loss, which is selected to validate the generalization ability. }

\begin{figure}[t]
\setlength{\abovecaptionskip}{0.12cm}
\setlength{\belowcaptionskip}{-0.55cm}
\centering
\includegraphics[width=1.0\columnwidth]{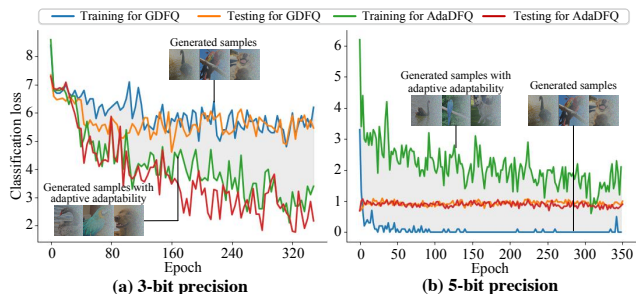}
\caption{Existing work, \emph{e.g.}, GDFQ \cite{xu2020generative} (the \textcolor[RGB]{30,118,180}{blue}), generally suffers from (a) underfitting issue (both training and testing loss are large) under 3-bit precision and (b) overfitting issue (training loss is small while testing loss is large) under 5-bit precision$^1$. Our AdaDFQ (the \textcolor[RGB]{44,160,44}{green}) generates the sample with adaptive adaptability to Q, yielding better generalization of Q with varied bit widths.
 The observations are from MobileNetV2 on ImageNet. }
\label{under_over}
\end{figure}

However, the original data may not be accessible due to privacy and security issues.
To this end, data-free quantization (DFQ) has come up to quantize models without the original data by synthesizing meaningful fake samples, where the quantized network (Q) is improved by distilling the knowledge from the pre-trained full-precision model (P) \cite{hinton2015distilling,qian2022switchable}.
Among the existing arts \cite{cai2020zeroq,zhang2021diversifying}, the increasing attention has recently transformed to the generative models \cite{xu2020generative,choi2021qimera,zhu2021autorecon}, which generally captures the distribution of the original data from P by utilizing the generative model as a generator (G), where P serves as the discriminator to guide the generation process \cite{xu2020generative}. To narrow the gap between the synthetic and real data, \cite{choi2021qimera} proposes to restore the decision boundaries via boundary supporting samples; while \cite{zhu2021autorecon} further optimizes the generator architecture for better fake samples via a neural architecture search method.
Nevertheless, it still suffers from a non-ignorable performance loss under various bit-width settings, stemming from the followings:
\begin{itemize}
    \item[(1)] Bearing the limited capacity of the generator, it is impossible for the generated sample with incomplete distribution to fully recover the original dataset, hence a natural question raises up as: whether the knowledge by the generated sample is \emph{informative} or not to benefit the learning process to Q? However, the generated sample by the prior arts customized for P, fail to benefit Q with varied bit-width settings (\emph{e.g.}, 3-bit or 5-bit precision), where only limited information from P can be exploited to recover Q.
    \item[(2)] In low-bit precision (\emph{e.g.}, 3 bit), Q suffers from a sharp accuracy drop upon P due to large quantization error, resulting into its poor learning performance. Following that, the generated sample by G may incur a large \emph{disagreement} between the predictions of P and Q, resulting the optimization loss into too large to converge, yielding an \emph{underfitting} issue; see Fig.\ref{under_over}(a).
    \item[(3)] In high-bit precision (\emph{e.g.}, 5 bit), Q possesses comparable recognition ability with P due to a small accuracy drop. Therefore, most of the generated samples by G, for which Q and P give similar predictions, may not benefit Q. However, trapped by the optimization loss, Q receives no improvement, resulting in an \emph{overfitting} issue; see Fig.\ref{under_over}(b).
\end{itemize}

\begin{table}[t]
\setlength{\abovecaptionskip}{0.32cm}
\caption{Comparison with the existing DFQ methods. Our AdaDFQ aims to generate the sample with adaptive adaptability to Q with varied bit widths, especially low-bit situation. }
\centering
\footnotesize
{\setlength\tabcolsep{1.6mm}
\begin{tabular}{c|c|c|c}
\toprule
 Method &\tabincell{c}{Generated  \\ sample type  }       &\tabincell{c}{Dependence  \\ on Q  }  &\tabincell{c}{Access to  \\ low-bit situation  }       \\
\hline\hline
GDFQ \cite{xu2020generative} &Reconstructed     &No  &No     \\
Qimera \cite{choi2021qimera} &Reconstructed     &No  &No     \\
ZAQ \cite{liu2021zero} &Adversarial     &Yes  &No     \\
\textcolor[RGB]{176,36,66}{AdaDFQ} &\cellcolor{mygreen}Adaptive adaptability     &\cellcolor{mygreen}Yes  &\cellcolor{mygreen}Yes     \\
\bottomrule
\end{tabular}
}
\label{sample_type}
\vspace{-2.4em}
\end{table}

\noindent The above conveys to us that the existing arts overlook the \emph{sample adaptability}, \emph{i.e.}, informative or not to Q, to Q with varied bit widths during the generation process from G, where Q is independent of the generation process; see Table \ref{sample_type}. Intuitively, (2) may crave the sample with large agreement between P and Q to avoid the underfitting issue; while for (3), it may be one with large disagreement to avoid the overfitting issue. Such fact promotes us to delve into the following questions: \emph{how to measure the sample adaptability to Q under varied bit-width scenarios? whether the largest adaptability is the best? how to generate the samples with adaptive adaptability to benefit Q’s generalization? }

To answer the above questions, we attempt to generate the samples with large adaptability to Q by taking Q into account during the generation process; see Table \ref{sample_type}.
Following that, \cite{qian2023rethinking} first reformulates the DFQ as a zero-sum game over sample adaptability between two players --- a generator and a quantized network, where one player’s reward is the other’s loss while their sum is zero.
Specifically, G generates the sample with large adaptability by enlarging the disagreement between P and Q, to benefit Q; while Q is calibrated to be improved by exploiting the sample with large adaptability. Such process of benefiting Q essentially decreases the sample adaptability, which is adversarial to increasing the sample adaptability for G. However, \cite{qian2023rethinking} fails to reveal the underlying \emph{underfitting} (\emph{overfitting}) issues incurred by the sample with \emph{largest} (\emph{lowest}) adaptability upon the zero-sum game process.

To address the above issues, we define two types of samples: disagreement (\emph{i.e.}, P can predict correctly but Q not) and agreement (\emph{i.e.}, P and Q have the same prediction) samples, to form the lower and upper boundaries to be balanced. The margin between two boundaries is optimized to adaptively regulate the sample adaptability under the constraint of over-and-under fitting issues, so as to generate the samples with \textit{adaptive} adaptability within these two boundaries to Q, with an \underline{Ada}ptive \underline{D}ata-\underline{F}ree \underline{Q}uantization (AdaDFQ) method. We further conduct the theoretical analysis on the generalization of Q, which reveals: the generated sample with the largest adaptability is NOT the best for Q's generalization; the knowledge carried by the generated sample should not only be informative to Q, but also related to the category and distribution information of the training data for P.
We remark that AdaDFQ optimizes the margin to generate the desirable samples upon the zero-sum game, aiming to address the \emph{over-and-under fitting} issues; while AdaSG \cite{qian2023rethinking} focuses primarily on the zero-sum game framework for DFQ, which serves as a \emph{special} case of AdaDFQ in spirit. The theoretical analysis and empirical studies validate the superiority of AdaDFQ to the state-of-the-arts.

\section{Adaptive Data-Free Quantization}
Conventional generative data-free quantization (DFQ) approaches \cite{xu2020generative,choi2021qimera,zhu2021autorecon} reconstruct the training data via a generator (G) by extracting the knowledge (\emph{i.e.}, the class distribution information about the original data) from a pre-trained full-precision network (P), to recover the performance of quantized network (Q) by the calibration operation.
However, we observe that Q is independent of the generation process by the existing arts and whether the knowledge carried by generated samples is informative or not to Q,  namely \emph{sample adaptability}, is crucial to Q's generalization.
Before shedding light on our method, we first focus on the sample adaptability to Q, followed by how to reformulate the DFQ  as a dynamic zero-sum game process over sample adaptability, as discussed in the next.

\subsection{How to Measure the Sample Adaptability to Quantized Network?}
To measure the sample adaptability to Q, we focus primarily on: 1) the dependence on Q; 2) the advantage of P over Q; and 3) the disagreement between P and Q.
Following \cite{qian2023rethinking}, we first define the disagreement sample below:

\noindent \textbf{Definition 1} (Disagreement Sample). \emph{Given a random noise vector $z\sim N(0,1)$ and an arbitrary one-hot label $y$, $G$ generates a sample $x=G(z|y)$. Then the logit outputs of $P$ and $Q$ are given as $z_p=P(x)$ and $z_q=Q(x)$, respectively. Suppose that $x$ can be correctly predicted by P, i.e., $arg max(z_p)=arg max(y)$, where $arg max(\cdot)$ returns the class index corresponding to maximum value. We determine $x$ to be the disagreement sample provided $arg max(z_p)\neq arg max(z_q)$.}

\noindent Thus the disagreement between P and Q is encoded by the following probability vector:
\begin{equation}
\begin{aligned}
&p_{ds}=softmax(z_p- z_q) \in \mathbb{R}^C,
\end{aligned}
\label{dist_disagree}
\end{equation}
where $C$ represents the number of class; $p_{ds}(c)=\frac{exp(z_p(c)-z_q(c))}{\sum_{j=1}^{C} exp(z_p(j)-z_q(j))}$ ($c\in\{1,2,...,C\}$) denotes the $c$-th entry of $p_{ds}$ as the probability that $x$ is identified as the disagreement sample of the $c$-th class.
Based on $p_{ds}$, the disagreement is calculated via the information entropy function $\mathcal{H}_{info}(\cdot)$, formulated as
\begin{small}
\begin{equation}
\mathcal{H}_{info}(p_{ds})=\sum^{C}_{c=1} p_{ds}(c)log\frac{1}{p_{ds}(c)}.
\label{metric}
\end{equation}
\end{small}
In view of the varied $C$ from different datasets, $\mathcal{H}_{info}(p_{ds})$ is further normalized as
\begin{small}
\begin{equation}
\begin{aligned}
\mathcal{H}_{nor}&=1-\mathcal{H}^{'}_{info}(p_{ds}) \in [0,1),
\end{aligned}
\label{measure_r}
\end{equation}
\end{small}
which can be exploited to measure the sample adaptability, where $\mathcal{H}^{'}_{info}(p_{ds})=\frac{\mathcal{H}_{info}(p_{ds})-min(\mathcal{H}_{info}(p_{ds}))}{max(\mathcal{H}_{info}(p_{ds}))-min(\mathcal{H}_{info}(p_{ds}))}$; the constant $max(\mathcal{H}_{info}(p_{ds}))=-\sum^{C}_{c=1}\frac{1}{C}log\frac{1}{C}$ represents the maximum value of $\mathcal{H}_{info}(p_{ds})$, where each element in $p_{ds}$ is $\frac{1}{C}$ (\emph{i.e.}, the same class probability), implying that Q perfectly aligns with P (\emph{i.e.}, $z_p=z_q$), while $min(\cdot)$ returns the minimum value of $\mathcal{H}_{info}(p_{ds})$ within a batch.

As inspired, we revisit the DFQ as: G generates the sample with large adaptability to Q, to be equivalent to maximizing $\mathcal{H}_{nor}$; upon the generated samples, Q is updated to recover the performance with decreasing the sample adaptability, which is equivalent to minimizing $\mathcal{H}_{nor}$, adversarial to maximizing $\mathcal{H}_{nor}$. Such fact is in line with the principle of zero-sum game$^2$\footnotetext{$^2$ For AdaDFQ, G aims to enlarge the disagreement, \emph{i.e.}, adaptability to Q, between P and Q, while Q decreases it by learning from P, where they cancel each other out, making the overall changing (summation) of the disagreement ($\mathcal{H}_{info}(p_{ds})$) \textit{close} to 0 (see Fig.\ref{ada_experiment} for such intuition), which is consistent to the intuition of zero-sum game.}\cite{v1928theorie}, as discussed in the next section.

\subsection{Zero-Sum Game over Sample Adaptability}
\label{dfq_game}
On top of \cite{qian2023rethinking}, we revisit the DFQ from a zero-sum game perspective over sample adaptability between two players --- a generator and a quantized network, as follows:
\begin{small}
\begin{equation}
\mathop{min}_{\theta_q\in \Theta_q} \mathop{max}_{\theta_g\in \Theta_g}  \mathcal{H}(\theta_g,\theta_q)=\mathop{min}_{\theta_q\in \Theta_q} \mathop{max}_{\theta_g\in \Theta_g}  \mathbb{E}_{z,y}[1-\mathcal{H}^{'}_{info}(p_{ds})],
\label{max_min}
\end{equation}
\end{small}
where $\theta_g\in \Theta_g$ and $\theta_q\in \Theta_q$ are the weight parameters of G and Q, respectively. Particularly, Eq.(\ref{max_min}) is alternatively optimized via gradient descent during each iteration: $\theta_q$ is fixed while $\theta_g$ is updated to generate the sample with \emph{large} adaptability  by \emph{maximizing} Eq.(\ref{max_min}); alternatively, $\theta_g$ is fixed while $\theta_q$ is updated to calibrate Q over the generated sample by \emph{minimizing} Eq.(\ref{max_min}).
The optimization process will reach a Nash equilibrium \cite{cardoso2019competing} $(\theta_g^*,\theta_q^*)$
when the following inequality
\begin{equation}
\mathcal{H}(\theta_g,\theta_q^*) \le \mathcal{H}(\theta_g^*,\theta_q^*) \le \mathcal{H}(\theta_g^*,\theta_q)
\label{}
\end{equation}
holds for all $\theta_g\in \Theta_g$ and $\theta_q\in \Theta_q$, where $\theta_g^*$ and $\theta_q^*$ are the parameters of G and Q under an equilibrium state. Maximizing Eq.(\ref{max_min}) is equivalent to maximizing $\mathcal{H}(\cdot,\cdot)$ throughout learning G to generate the sample with largest adaptability (\emph{i.e.}, smallest $\mathcal{H}^{'}_{info}(p_{ds})$). 

However, such fact may incur:

\noindent \textbf{Underfitting issue:} the knowledge carried by the generated samples from G (\emph{e.g.}, \textcolor[RGB]{237,125,49}{$\bigcirc$} in Fig.\ref{sample_dist}), exhibits excessive information with large adaptability, implying a large disagreement between P and Q, while Q (especially for Q with low bit width) has no sufficient ability to learn informative knowledge from P. Evidently, the sample with the largest adaptability (\emph{i.e.}, smallest $\mathcal{H}^{'}_{info}(p_{ds})$) is not the best. For such case, Q is calibrated by minimizing Eq.(\ref{max_min}) over such samples to incur the underfitting, which, in turn, encourages G to generate the samples with lower adaptability (\emph{i.e.}, larger $\mathcal{H}^{'}_{info}(p_{ds})$) by alternatively maximizing Eq.(\ref{max_min}). However, encouraging the sample with lowest adaptability (\emph{i.e.}, largest $\mathcal{H}^{'}_{info}(p_{ds})$) may lead to:

\noindent \textbf{Overfitting issue:} the knowledge carried by the generated samples (\emph{e.g.}, \textcolor[RGB]{0,176,240}{$\diamondsuit$} in Fig.\ref{sample_dist}) delivers limited information with small adaptability to yield a large agreement between P and Q. For such case, it may not be informative to calibrate Q (especially for Q with high bit width) by minimizing Eq.(\ref{max_min}), which alternatively encourages G to generate the samples with the larger adaptability by maximizing Eq.(\ref{max_min}).

\subsection{Refining the Maximization of Eq.(\ref{max_min}): Generating the Sample with Adaptive Adaptability}
\label{refining_eq5}
The above facts indicate that the sample with either largest or lowest adaptability generated by maximizing Eq.(\ref{max_min}) is not necessarily the best, incurring over-and-under fitting issues, which fail to be resolved by the above disagreement sample (Definition 1) since it focuses on the larger adaptability (\emph{i.e.}, smaller $\mathcal{H}^{'}_{info}(p_{ds})$).
To address the issues, we refine the maximization of Eq.(\ref{max_min}) during the zero-sum game by proposing to balance disagreement sample with agreement sample, as discussed in the next.

\subsubsection{Balancing Disagreement Sample with Agreement Sample}
\label{balance_ds_as}
As per Definition 1, the category information from P is crucial to establish the dependence of generated sample on Q.  Thereby, to generate the disagreement sample with adaptive adaptability, we exploit the category information to guide G's optimization. Given the class label $y$, it is expected that the generated sample is identified as disagreement sample with the same label $y$. Accordingly, we present to match $p_{ds}$ and $y$ via the Cross-Entropy loss $\mathcal{H}_{CE}(.,.)$ below:
\begin{equation}
\mathcal{L}_{ds}=\mathbb{E}_{z,y}[\mathcal{H}_{CE}(p_{ds},y)].
\label{ds_ce}
\end{equation}
Eq.(\ref{ds_ce}) encourages G to generate the disagreement sample that P can predict correctly but Q fails. However, the disagreement sample tends to yield smaller $\mathcal{H}^{'}_{info}(p_{ds})$, which mitigates the overfitting while the underfitting issue is still available.
To remedy such issue, we further define the agreement sample to weaken the effect of disagreement sample on reducing $\mathcal{H}^{'}_{info}(p_{ds})$.

\noindent \textbf{Definition 2} (Agreement Sample). \emph{Based on Definition 1, we determine the generated sample $x$ to be the agreement sample if $arg max(z_p)=arg max(z_q)$.}

\begin{figure}[t]
\centering
\setlength{\abovecaptionskip}{0.02cm}
\setlength{\belowcaptionskip}{-0.55cm}
\includegraphics[width=0.75\columnwidth]{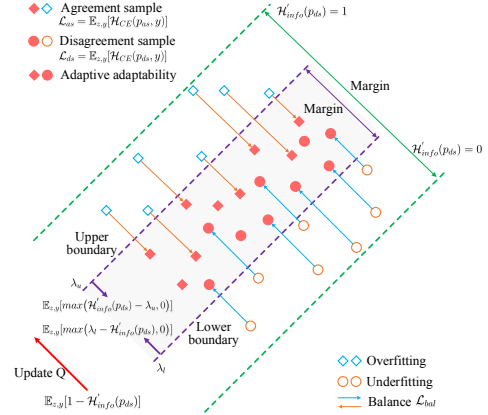}
\caption{Illustration of generating the sample with adaptive adaptability to address the over-and-under fitting issues.
}
\label{sample_dist}
\end{figure}

\noindent Thus, similar to $p_{ds}$ in Eq.(\ref{dist_disagree}), the agreement between P and Q is encoded via the probability vector
\begin{equation}
p_{as}=softmax(z_p+z_q) \in \mathbb{R}^C,
\end{equation}
where $p_{as}(c)=\frac{exp(z_p(c)+z_q(c))}{\sum_{j=1}^{C} exp(z_p(j)+z_q(j))}$ is the $c$-th entry of $p_{as}$ as the probability that $x$ is the agreement sample of the $c$-th class. Following Eq.(\ref{ds_ce}), $p_{as}$ and $y$ are matched via the following loss function:
\begin{equation}
\mathcal{L}_{as}=\mathbb{E}_{z,y}[\mathcal{H}_{CE}(p_{as},y)].
\label{las}
\end{equation}
Eq.(\ref{las}) encourages G to generate the agreement sample that both P and Q can correctly predict, yielding larger $\mathcal{H}^{'}_{info}(p_{ds})$.
Intuitively, the agreement sample is capable of balancing disagreement sample to avoid too small or large $\mathcal{H}^{'}_{info}(p_{ds})$ for adaptive adaptability, formulated as
\begin{small}
\begin{equation}
\begin{aligned}
&\mathcal{L}_{bal}=\alpha_{ds} \mathcal{L}_{ds} + \alpha_{as} \mathcal{L}_{as},
\label{L_reg}
\end{aligned}
\end{equation}
\end{small}
where $\alpha_{ds}$ and $\alpha_{as}$ are used to facilitate the balance between $\mathcal{L}_{ds}$ and $\mathcal{L}_{as}$ (see the parameter study). Eq.(\ref{L_reg}) denotes the loss between $p_{ds}$, $p_{as}$ and $y$ to be minimized during the training phase.
From this perspective, $\mathcal{L}_{ds}$ attempts to generate the disagreement samples to enlarge the gap between P and Q , so as to reduce $\mathcal{H}^{'}_{info}(p_{ds})$ for agreement samples (see \textcolor[RGB]{237,125,49}{$\leftarrow$} in Fig.\ref{sample_dist}); to be analogous, $\mathcal{L}_{as}$ enlarges $\mathcal{H}^{'}_{info}(p_{ds})$ for disagreement samples (see \textcolor[RGB]{0,176,240}{$\rightarrow$} in Fig.\ref{sample_dist}).

Intuitively, $\mathcal{L}_{bal}$ endows the generated sample with adaptive adaptability, \emph{i.e.}, neither too large nor small $\mathcal{H}^{'}_{info}(p_{ds})$ for the generated samples throughout the balance between disagreement and agreement samples. In other words, we need to study how to control $\mathcal{H}^{'}_{info}(p_{ds})$ \emph{within a desirable range} via the balance process; cored on that, it establishes the implicit lower and upper boundary corresponding to the disagreement and agreement samples (see Fig.\ref{sample_dist}), hence the above curse is equivalent to confirming the margin between such lower-and-upper boundary. To this end, we propose to optimize the margin between these two boundaries, as discussed in the next.

\subsubsection{Optimizing the Margin Between two Boundaries}
\label{lu_boundary}
Formally, along with $\mathcal{L}_{bal}$, the maximization objective of Eq.(\ref{max_min}) is imposed with the desirable bound constraints for $\mathcal{H}^{'}_{info}(p_{ds})$ (see Fig.\ref{sample_dist}), and formulated as
\begin{small}
\begin{equation}
\begin{aligned}
&\mathop{max}_{\theta_g\in \Theta_g} \  \mathcal{H}(\theta_g,\theta_q) - \beta \mathcal{L}_{bal},  \\
&\text{subject to} \quad  \lambda_l < \mathcal{H}_{info}^{'}(p_{ds}) < \lambda_u,
\label{extra_constraint}
\end{aligned}
\end{equation}
\end{small}
where $\beta$ is utilized to balance the agreement and disagreement sample. $\lambda_l$ and $\lambda_u$ denote the lower and upper bound of $\mathcal{H}^{'}_{info}(p_{ds})$, such that $0 \le \lambda_l < \lambda_u \le 1$. where the goal is to ensure the balance process to generate the sample with adaptive adaptability.
$\lambda_l$ serves to prevent G from generating the sample with too large adaptability (\emph{i.e.}, too small $\mathcal{H}^{'}_{info}(p_{ds})$) via maximizing Eq.(\ref{max_min}), where Q is calibrated by minimizing Eq.(\ref{max_min}) to incur \emph{underfitting}. By contrast,  $\lambda_u$ aims to avoid the sample with too small adaptability (\emph{i.e.}, too large $\mathcal{H}^{'}_{info}(p_{ds})$), which is not informative to calibrate Q by minimizing Eq.(\ref{max_min}), causing \emph{overfitting} issue.

\noindent \textbf{Discussion on $\lambda_l$} and \textbf{$\lambda_u$}.
The crucial issue is how to specify $\lambda_l$ and $\lambda_u$? First, we investigate the ideal condition where Q can be well calibrated over varied generated samples from G, it is ideally expected that each generated sample possesses an optimal pairwise of $\lambda_l$ and $\lambda_u$, to help calibrate Q by minimizing Eq.(\ref{max_min}), which, unfortunately, is infeasible for individual sample since a large number of generated samples within a batch are generated from G to calibrate Q, only single sample makes no sense; for each batch, the adaptability of each single sample is merely affected by its preceding generated sample, and independent of all other generated samples from G; worse still, the dependencies between varied batches are quite uncertain, since Q varies after being calibrated for each batch. As inspired, instead of trapping into the adaptive rule to each individual sample, we propose to acquire the range of $\lambda_l$ and $\lambda_u$ for all generated samples.
Particularly, we uniformly select the values of $\lambda_l$ and $\lambda_u$ for all samples from $[0,1]$ (see the extensive parameter study), to answer ``how much" to the balancing process (Eq.(\ref{L_reg})), such that the following demands meet. That is, avoiding too small $\lambda_l$ or large $\lambda_u$, to address the over-and-under fitting issues, so as to ensure G to well generate the samples with adaptive adaptability.

The above fact discussed the adaptability of single sample to Q, while the calibration process for Q involves the generated samples from a batch. As inspired, we exploit the batch normalization statistics (BNS) information \cite{xu2020generative,choi2021qimera} about the training data for P, which is achieved by
\begin{small}
\begin{equation}
\mathcal{L}_{BNS}=\sum^M_{m=1}( ||\mu^g_m-\mu_m||^2_2 + ||\sigma^g_m-\sigma_m||^2_2 ),
\label{bns}
\end{equation}
\end{small}
where $\mu^g_m / \sigma^g_m$ and $\mu_m / \sigma_m$ denote the mean/variance of the generated samples' distribution within a batch and the corresponding mean/variance parameters for P at the $m$-th BN layer of the total $M$ layers. Eq.(\ref{bns}) encodes the loss between them to be minimized during the training.
To this end, the maximization objective of Eq.(\ref{max_min}) is refined as
\begin{small}
\begin{equation}
\begin{aligned}
\mathop{max}_{\theta_g\in \Theta_g} & \mathbb{E}_{z,y}[-max\big(\lambda_l-\mathcal{H}_{info}^{'}(p_{ds}),0\big)]+ \\
& \mathbb{E}_{z,y}[-max\big(\mathcal{H}_{info}^{'}(p_{ds})-\lambda_u,0\big)] - \beta \mathcal{L}_{bal} - \gamma \mathcal{L}_{BNS},
\label{rewritten_eq10}
\end{aligned}
\end{equation}
\end{small}
where the first two terms aim to optimize the margin between $\lambda_l$ and $\lambda_u$ via the hinge loss \cite{lim2017geometric}. $max(\cdot, \cdot)$ returns the maximum. $\beta$ and $\gamma$ are balance parameters, where, the minus ``$-$" for $\mathcal{L}_{bal}$ and $\mathcal{L}_{BNS}$ is noted be minimized during the optimization as aforementioned.
With Eq.(\ref{rewritten_eq10}), the minimization objective of Eq.(\ref{max_min}) holds to calibrate Q for performance recovery, yielding:
\begin{small}
\begin{equation}
\mathop{min}_{\theta_q\in \Theta_q} \ \mathbb{E}_{z,y}[1-\mathcal{H}^{'}_{info}(p_{ds})].
\label{min_term}
\end{equation}
\end{small}
By alternatively optimizing Eq.(\ref{rewritten_eq10}) and (\ref{min_term}) during a zero-sum game, the samples with adaptive adaptability can be generated by G to maximally recover the performance of Q until reaching a Nash equilibrium.

\subsection{Theoretical Analysis: Why do the Boundaries Improve Q's Generalization?}
\label{generalization_analysis}
One may wonder why AdaDFQ can generate the sample with adaptive adaptability. We answer the question by studying the generalization of Q trained on the generated samples from a statistic view.
According to the VC theory \cite{vapnik1999nature,lopez2015unifying,mirzadeh2020improved}, the classification error of a quantized network (Q) learning from the ground truth label (y) on the generated samples by G can be decomposed as
\begin{small}
\begin{equation}
\begin{aligned}
&R(f_q)-R(f_r) \le O(\frac{|\mathcal{F}_q|_C}{n^{\alpha_{qr}}}) + \varepsilon_{qr},
\end{aligned}
\label{}
\end{equation}
\end{small}
where $R(\cdot)$ is the error of a specific function. $f_q \in \mathcal{F}_q$ is the quantized network (Q) function and $f_r \in \mathcal{F}_r$ is the ground truth label ($y$) function. $\varepsilon_{qr}$ denotes the approximation error of the quantized network function class $\mathcal{F}_q$ (considering all possible acquired $f_q$ by optimizing Q) w.r.t. $f_r \in \mathcal{F}_r$ (a fixed learning target) on the generated samples during the training phase. $\alpha_{qr}$ denotes the learning rate for the given generated samples, particularly: when $\alpha_{qr}$ approaches $\frac{1}{2}$ (a slow rate) for the generated samples with the excessive information to Q; while approaching $1$ (a fast rate) for the generated samples with the limited information to Q. $O(\cdot)$ represents the estimation error of a network function over the real data during the testing phase. $|\cdot|_C$ measures the capacity of a function class and $n$ is the number of the generated samples from G.
Similarly, let $f_p \in \mathcal{F}_p$ be the full-precision network (P) function, then
\begin{small}
\begin{equation}
\begin{aligned}
&R(f_p)-R(f_r) \le O(\frac{|\mathcal{F}_p|_C}{n^{\alpha_{pr}}}) + \varepsilon_{pr},
\end{aligned}
\label{teacher_real}
\end{equation}
\end{small}
where  $\alpha_{pr}$ is related to the learning rate of P upon the ground truth label ($y$); $\varepsilon_{pr}$ denotes the approximation error of the full-precision network (P) function class $\mathcal{F}_p$ w.r.t. $f_r \in \mathcal{F}_r$.
For data-free setting, Q is required to learn from P with the generated samples, which yields the following:
\begin{small}
\begin{equation}
\begin{aligned}
&R(f_q)-R(f_p) \le O(\frac{|\mathcal{F}_q|_C}{n^{\alpha_{qp}}}) + \varepsilon_{qp},
\end{aligned}
\label{student_teacher}
\end{equation}
\end{small}
where $\alpha_{qp}$ is related to the learning rate of Q upon P, while $\varepsilon_{qp}$ is the approximation error of the quantized network (Q) function class $\mathcal{F}_q$ w.r.t. $f_p \in \mathcal{F}_p$.
For such case, to study the classification error of Q learning from the ground truth label $y$ on the generated samples, we combine Eq.(\ref{teacher_real}) and (\ref{student_teacher}) \cite{lopez2015unifying,mirzadeh2020improved}, leading to
\begin{footnotesize}
\begin{equation}
\begin{aligned}
R(f_q)-R(f_r)  & = R(f_q)-R(f_p) + R(f_p)-R(f_r) \\
&\le O(\frac{|\mathcal{F}_q|_C}{n^{\alpha_{qp}}}) + \varepsilon_{qp} + O(\frac{|\mathcal{F}_p|_C}{n^{\alpha_{pr}}}) + \varepsilon_{pr} \\
& = \underbrace{ O(\frac{|\mathcal{F}_q|_C}{n^{\alpha_{qp}}}) + O(\frac{|\mathcal{F}_p|_C}{n^{\alpha_{pr}}}) }_{\text{Estimation error}} + \underbrace{ \varepsilon_{qp} + \varepsilon_{pr}. }_{\text{Approximation error}}
\end{aligned}
\label{upper_qr}
\end{equation}
\end{footnotesize}
Evidently, Eq.(\ref{upper_qr}) shows that, to benefit the generalization of Q, both estimation and approximation error should be reduced, so that a tighter upper bound can be captured, where we claim that to be consistent with the insights on the optimization of Eq.(\ref{rewritten_eq10}): \emph{first}, as aforementioned, the reasons for the overflow of generalization error stem from:
1) the generated sample with too large $\mathcal{H}^{'}_{info}(p_{ds})$ is not informative to calibrate Q, \emph{i.e.}, the small  $\varepsilon_{qp}$ during the training stage and large $O(\frac{|\mathcal{F}_q|_C}{n^{\alpha_{qp}}})$ during the testing stage, leading to \emph{overfitting} issue for Q;
2) for the generated sample with too small $\mathcal{H}^{'}_{info}(p_{ds})$, Q has no sufficient ability to learn informative knowledge from P, \emph{i.e.}, both $\varepsilon_{qp}$ during the training stage and $O(\frac{|\mathcal{F}_q|_C}{n^{\alpha_{qp}}})$ during the testing stage are large, leading to \emph{underfitting} issue for Q. Based on that, $\lambda_l$ and $\lambda_u$ in Eq.(\ref{extra_constraint}) along with the balance process ($\mathcal{L}_{bal}$) in Eq.(\ref{L_reg}) aim to avoid too large or small $\varepsilon_{qp}$ by optimizing the margin, thus the estimation error $O(\frac{|\mathcal{F}_q|_C}{n^{\alpha_{qp}}})$ can be reduced,  \emph{i.e.}, overcoming the \emph{over-and-under fitting} issues.
\emph{Second}, the above fact discussed the adaptability of single sample to Q, as aforementioned, BNS distribution information (Eq.(\ref{bns})) of the training data extracted from P further facilitates calibrating Q for better generalization via the generated samples within a batch, which ensures the generated samples to be informative to P, \emph{i.e.}, decreasing $O(\frac{|\mathcal{F}_p|_C}{n^{\alpha_{pr}}}) + \varepsilon_{pr}$ in Eq.(\ref{teacher_real}) or (\ref{upper_qr}), along with the category information from P regarding the disagreement and agreement samples.

Based on the above, the sample with adaptive adaptability to Q can be generated by optimizing Eq.(\ref{rewritten_eq10}), reducing both estimation and approximation error (obtaining a tighter upper bound) in Eq.(\ref{upper_qr}), so as to generate the samples by G with adaptive adaptability
which is beneficial to calibrating Q for better generalization by optimizing Eq.(\ref{min_term}), where they are alternatively optimized during a zero-sum game until reaching a Nash equilibrium.

\section{Experiment}
\subsection{Experimental Settings and Details}
We validate AdaDFQ over three typical image classification datasets, including: \textbf{CIFAR-10} and \textbf{CIFAR-100} \cite{krizhevsky2009learning} contain 10 and 100 classes of images, which are split into 50K training images and 10K testing images; \textbf{ImageNet} (\textbf{ILSVRC2012}) \cite{russakovsky2015imagenet} consists of 1.2M samples for training and 50k samples for validation with 1000 categories. For data-free setting, only validation sets are adopted to evaluate the performance of the quantized models (Q).
We quantize pre-trained full-precision networks (P) including ResNet-20 for CIFAR, and ResNet-18, ResNet-50 and MobileNetV2 for ImageNet, via the following quantizer to yield Q:

\noindent \textbf{Quantizer.}
Following \cite{xu2020generative,choi2021qimera}, we quantize both full-precision (float32) weights and activations into $n$-bit precision by a symmetric linear quantization method as \cite{jacob2018quantization}:
\begin{small}
\begin{equation}
\theta_q=round \Big ((2^n-1) * \frac{\theta-\theta_{min}}{\theta_{max}-\theta_{min}}-2^{n-1} \Big),
\end{equation}
\end{small}
where $\theta$ and $\theta_q$ are the full-precision and quantized value. $round(\cdot)$ returns the nearest integer value to the input. $\theta_{min}$ and $\theta_{max}$ are the minimum and maximum of $\theta$.

For \emph{generation} process, we construct the architecture of the generator G following ACGAN \cite{odena2017conditional}, which is trained via Eq.(\ref{rewritten_eq10}) using Adam \cite{kingma2014adam} as an optimizer with a momentum of 0.9 and a learning rate of 1e-3.
For \emph{calibration} process, Q is optimized by minimizing Eq.(\ref{min_term}), where SGD with Nesterov \cite{nesterov1983method} is adopted as an optimizer with a momentum of 0.9 and weight decay of 1e-4. For CIFAR, the learning rate is initialized to 1e-4 and decayed by 0.1 for every 100 epochs; while it is 1e-5 (1e-4 for ResNet-50) and divided by 10 at epoch 350 on ImageNet. G and Q are alternatively trained for 400 epochs. The batch size is set to 16. For hyper-parameters,  $\alpha_{ds}$ and $\alpha_{as}$ in Eq.(\ref{L_reg}); $\lambda_l$, $\lambda_u$, $\beta$ and $\gamma$ in Eq.(\ref{rewritten_eq10}) are empirically set to 0.2, 0.1, 0.1, 0.8, 1 and 1 (\emph{see our supplementary material for more parameter studies}).
All experiments are implemented with pytorch \cite{paszke2019pytorch} via the code of GDFQ \cite{xu2020generative} and run on an NVIDIA GeForce GTX 1080 Ti GPU and an Intel(R) Core(TM) i7-6950X CPU @ 3.00GHz.

To evaluate AdaDFQ, we offer practical insights into ``why" AdaDFQ works, including the comparisons with the state-of-the-arts, ablation studies apart from visual analysis.

\begin{figure}[t]
\centering
\setlength{\abovecaptionskip}{0.01cm}
\setlength{\belowcaptionskip}{-0.25cm}
\includegraphics[width=1.0\columnwidth]{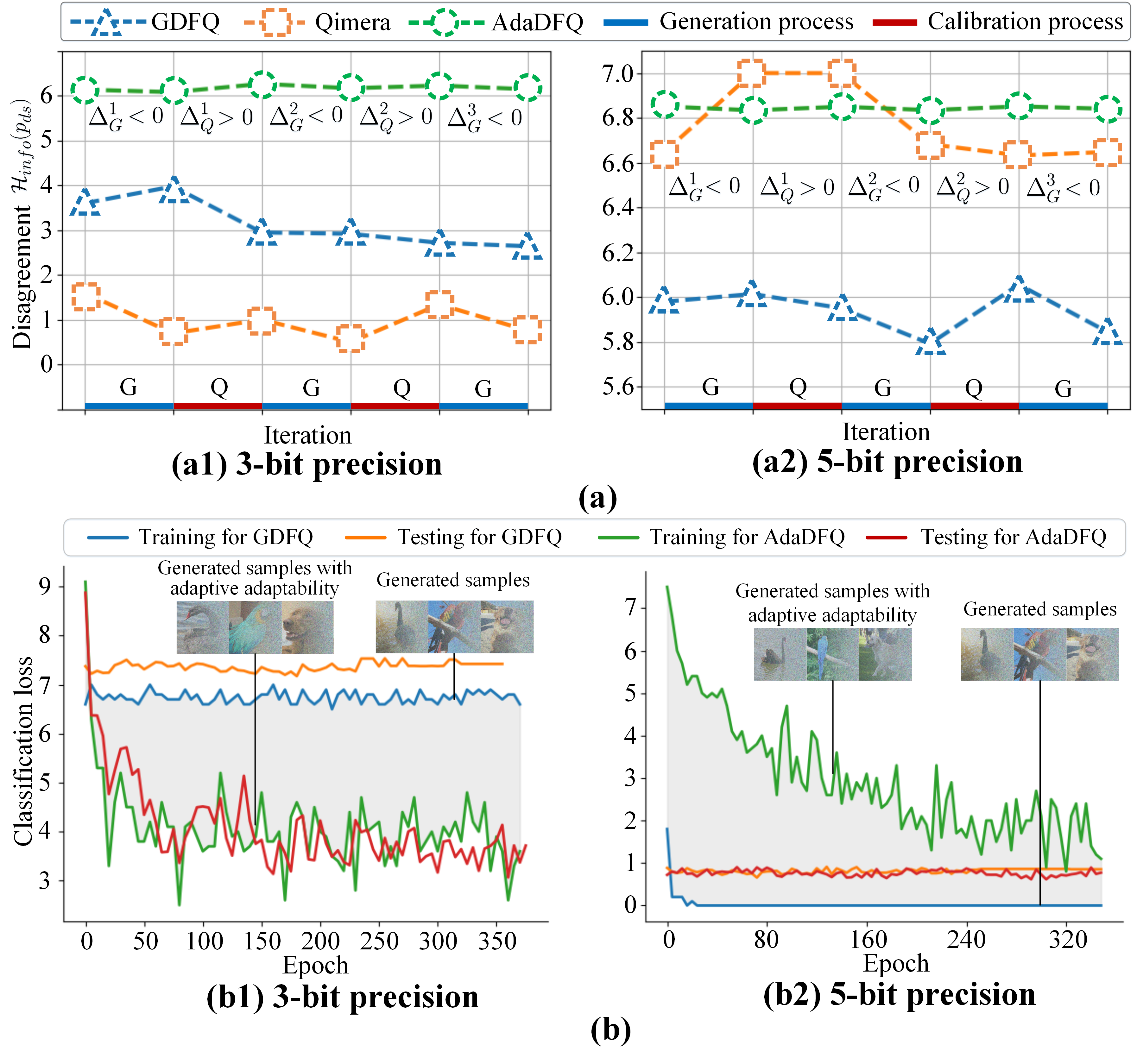}
\caption{Illustration of AdaDFQ on generating the samples with adaptive adaptability to Q under 3-bit and 5-bit precision. {(a)} Disagreement between P and Q during the generation (\textcolor[RGB]{0,112,192}{\textbf{---}}) and calibration (\textcolor[RGB]{192,0,0}{\textbf{---}}) process. $\Delta_Q^i>0$ and $\Delta_G^i<0$ denote the positive and negative gain of the disagreement, \emph{i.e.,} $\mathcal{H}_{info}(p_{ds})$, at $i$-th iteration of zero-sum game for the sample generation from G and calibration to Q. {(b)} Classification loss for Q during the training and testing phases. }
\label{ada_experiment}
\end{figure}

\begin{table*}
\setlength{\abovecaptionskip}{0.02cm}
\caption{Accuracy (\%) comparison with the state-of-the-arts on CIFAR-10, CIFAR-100 and ImageNet. $\dag$: the results implemented by author-provided code. -: no results are reported. \emph{n}w\emph{n}a indicates the weights and activations are quantized to \emph{n}-bit. The best results are reported with \textbf{boldface}.}
\label{sota}
\centering
\scriptsize
{\setlength\tabcolsep{1.2mm}
\begin{tabular}{c||cc||  c c c cccc  ||c}
\toprule
\textbf{Dataset} &\tabincell{c}{\textbf{Model}  \\ (Full precision)  }     & \textbf{Bit width}  &\tabincell{c}{\textbf{\textcolor[RGB]{182,130,75}{ZAQ}} \cite{liu2021zero} \\ (CVPR 2021)} &\tabincell{c}{\textbf{\textcolor[RGB]{112,48,160}{IntraQ}} \cite{zhong2022intraq}  \\ (CVPR 2022)} &\tabincell{c}{\textbf{\textcolor[RGB]{239,134,51}{ARC+AIT}} \cite{choi2022s}  \\ (CVPR 2022)}  &\tabincell{c}{\textbf{\textcolor[RGB]{0,176,240}{GDFQ}} \cite{xu2020generative} \\ (ECCV 2020)}    &\tabincell{c}{\textbf{\textcolor[RGB]{0,176,240}{ARC}} \cite{zhu2021autorecon} \\ (IJCAI 2021)}    &\tabincell{c}{\textbf{\textcolor[RGB]{0,176,240}{Qimera}} \cite{choi2021qimera} \\ (NeurIPS 2021)}  &\tabincell{c}{\textbf{\textcolor[RGB]{0,195,195}{AdaSG}} \cite{qian2023rethinking} \\ (AAAI 2023)}    &\tabincell{c}{ \textbf{\textcolor[RGB]{176,36,66}{AdaDFQ}} \\ \textbf{\textcolor[RGB]{176,36,66}{(Ours)}}  }   \\
\hline\hline
\multirow{3}{*}{\textbf{CIFAR-10}}
&                       &\emph{3}w\emph{3}a     &-      &77.07 &-  &\ 75.11$^{\dag}$     &-     &\ 74.43$^{\dag}$   &84.14         &\textbf{84.89}   \\
&\textbf{\emph{ResNet-20}}      &\emph{4}w\emph{4}a     &\cellcolor{mygreen}92.13 &\cellcolor{mygreen}91.49 &\cellcolor{mygreen}90.49     &\cellcolor{mygreen}90.11     &\cellcolor{mygreen}88.55     &\cellcolor{mygreen}91.26  &\cellcolor{mygreen}92.10  &\cellcolor{mygreen}\textbf{92.31}  \\
&(93.89)            &\emph{5}w\emph{5}a     &93.36 &-  &92.98   &93.38     &92.88     &93.46  &93.76  &\textbf{93.81}   \\
\hline\hline
\multirow{3}{*}{\textbf{CIFAR-100}}
&                       &\emph{3}w\emph{3}a     &-    &48.25 &41.34  &\ 47.61$^{\dag}$      &40.15    &\ 46.13$^{\dag}$     &\textbf{52.76}      &{52.74}   \\
&\textbf{\emph{ResNet-20}}      &\emph{4}w\emph{4}a     &\cellcolor{mygreen}60.42 &\cellcolor{mygreen}64.98 &\cellcolor{mygreen}61.05    &\cellcolor{mygreen}63.75      &\cellcolor{mygreen}62.76    &\cellcolor{mygreen}65.10   &\cellcolor{mygreen}66.42   &\cellcolor{mygreen}\textbf{66.81}  \\
&(70.33)            &\emph{5}w\emph{5}a     &68.70 &-  &68.40   &67.52      &68.40    &69.02  &69.42  &\textbf{69.93}   \\

\hline\hline
\multirow{3}{*}{\textbf{}}
&                       &\emph{3}w\emph{3}a    &-      &- &-  &\ 20.23$^{\dag}$   &23.37       &1.17$^{\dag}$  &37.04    &\textbf{38.10}   \\
&\textbf{\emph{ResNet-18}}      &\emph{4}w\emph{4}a    &\cellcolor{mygreen}52.64 &\cellcolor{mygreen}66.47 &\cellcolor{mygreen}65.73     &\cellcolor{mygreen}60.60   &\cellcolor{mygreen}61.32       &\cellcolor{mygreen}63.84  &\cellcolor{mygreen}66.50   &\cellcolor{mygreen}\textbf{66.53}  \\
&(71.47)            &\emph{5}w\emph{5}a    &64.54 &69.94  &70.28    &68.49   &68.88       &69.29   &\textbf{70.29}    &\textbf{70.29}   \\

\cline{2-11}
\multirow{3}{*}{\textbf{ImageNet}}
&                       &\emph{3}w\emph{3}a    &-     &- &- &1.46$^{\dag}$     &14.30    &-    &26.90            &\textbf{28.99}   \\
&\textbf{\emph{MobileNetV2}}  &\emph{4}w\emph{4}a    &\cellcolor{mygreen}0.10  &\cellcolor{mygreen}65.10 &\cellcolor{mygreen}\textbf{66.47}   &\cellcolor{mygreen}59.43    &\cellcolor{mygreen}60.13    &\cellcolor{mygreen}61.62    &\cellcolor{mygreen}65.15   &\cellcolor{mygreen}65.41  \\
&(73.03)           &\emph{5}w\emph{5}a    &62.35 &71.28 &\textbf{71.96}    &68.11   &68.40   &70.45  &71.61    &{71.61}   \\

\cline{2-11}
\multirow{3}{*}{\textbf{}}
&                       &\emph{3}w\emph{3}a      &-    &-  &- &0.31$^{\dag}$      &1.63    &-    &16.98          &\textbf{17.63}   \\
&\textbf{\emph{ResNet-50}}      &\emph{4}w\emph{4}a      &\cellcolor{mygreen}53.02  &\cellcolor{mygreen}-  &\cellcolor{mygreen}68.27 &\cellcolor{mygreen}54.16    &\cellcolor{mygreen}64.37   &\cellcolor{mygreen}66.25  &\cellcolor{mygreen}\textbf{68.58}    &\cellcolor{mygreen}{68.38}  \\
&(77.73)            &\emph{5}w\emph{5}a      &73.38  &-  &76.00  &71.63    &74.13   &75.32   &\textbf{76.03}     &\textbf{76.03}   \\
\bottomrule
\end{tabular}
}
\vspace{-1.3em}
\end{table*}

\subsection{Why does AdaDFQ Work?}
We verify the core idea of AdaDFQ --- optimizing the margin to generate the sample with adaptive adaptability for better Q's generalization under varied bit widths. We perform the experiments with ResNet-18 (Fig.\ref{ada_experiment}(a)) and ResNet-50 (Fig.\ref{ada_experiment}(b)) serving as both P and Q on ImageNet.
Fig.\ref{ada_experiment}(a) illustrates that, compared to GDFQ \cite{xu2020generative} and Qimera \cite{choi2021qimera}, the disagreement (computed by Eq.(\ref{metric})) between P and Q for AdaDFQ performs stably within a small range, \emph{i.e.}, the overall changing summation ($\Delta_G^i + \Delta_Q^i$) of the disagreement is \emph{close} to 0, following the principle of zero-sum game (Sec.\ref{dfq_game}),  which confirms that the generated sample with adaptive adaptability by AdaDFQ is fully exploited to benefit Q, and the lower and upper bound constraints ($\lambda_l$ and $\lambda_u$ in Eq.(\ref{rewritten_eq10})) avoid the generated sample with too large or small adaptability, which results in ridiculously large disagreement or agreement.
Fig.\ref{ada_experiment}(b) reveals that AdaDFQ achieves better Q's generalization with 3-bit and 5-bit precision unlike GDFQ, where the generated sample with adaptive adaptability succeeds in overcoming the \emph{underfitting} (both training and testing loss are large) and \emph{overfitting} (small training loss but large testing loss), confirming the analysis in Sec.\ref{generalization_analysis}.

\begin{table}[t]
\setlength{\abovecaptionskip}{0.02cm}
\caption{Ablation study about varied components of AdaDFQ on ImageNet. \emph{n}w\emph{n}a indicates the weights and activations are quantized to \emph{n}-bit. The best results are reported with \textbf{boldface}.}
\centering
\scriptsize
\begin{tabular}{c||cccc||c||c}
\toprule
\tabincell{c}{\textbf{Model}  \\ (Full-precision)  } &$\mathcal{L}_{ds}$ &$\mathcal{L}_{as}$     &$\lambda_l$, $\lambda_u$  &$\mathcal{L}_{BNS}$    &\emph{3}w\emph{3}a  &\emph{5}w\emph{5}a   \\
\hline\hline

\multirow{6}{*}{\tabincell{c}{\textbf{\emph{ResNet-18}} \\ (71.47)}}
& &\ding{51}      &\ding{51}           &\ding{51}      &19.40     &70.03   \\

&\cellcolor{mygreen}\ding{51}&\cellcolor{mygreen}      &\cellcolor{mygreen}\ding{51}           &\cellcolor{mygreen}\ding{51}       &\cellcolor{mygreen}31.14     &\cellcolor{mygreen}69.77  \\

&&      &\ding{51}           &\ding{51}                                &18.53     &66.27   \\

&\cellcolor{mygreen}\ding{51}&\cellcolor{mygreen}\ding{51}      &\cellcolor{mygreen}           &\cellcolor{mygreen}\ding{51}                  &\cellcolor{mygreen}32.13     &\cellcolor{mygreen}70.06  \\

&\ding{51}&\ding{51}      &\ding{51}           &                  &20.99     &67.80  \\

&\cellcolor{mygreen}\ding{51}&\cellcolor{mygreen}\ding{51}      &\cellcolor{mygreen}\ding{51}           &\cellcolor{mygreen}\ding{51}    &\cellcolor{mygreen}\textbf{38.10}     &\cellcolor{mygreen}\textbf{70.29}  \\

\bottomrule
\end{tabular}
\label{ablation}
\vspace{-1.0em}
\end{table}

\subsection{Comparison with State-of-the-arts}
To verify the superiority of AdaDFG, we compare it with typical DFQ methods, \emph{i.e.}, \textcolor[RGB]{0,176,240}{GDFQ} \cite{xu2020generative}, \textcolor[RGB]{0,176,240}{ARC} \cite{zhu2021autorecon} and \textcolor[RGB]{0,176,240}{Qimera} \cite{choi2021qimera}: reconstructing the original data from P; \textcolor[RGB]{182,130,75}{ZAQ} \cite{liu2021zero}  focuses primarily on the adversarial sample generation rather than adversarial game process for AdaDFG; \textcolor[RGB]{112,48,160}{IntraQ} \cite{zhong2022intraq} optimizes the noise to obtain fake sample without a generator; \textcolor[RGB]{239,134,51}{AIT} \cite{choi2022s} improves the loss function and gradients for ARC to generate better sample, denoted as \textcolor[RGB]{239,134,51}{ARC+AIT};
\textcolor[RGB]{0,195,195}{AdaSG} \cite{qian2023rethinking} focuses on the zero-sum game framework, serving as a special case of AdaDFQ.

Table \ref{sota} summarizes our following findings: 1) AdaDFQ obtains a significant and consistent accuracy gain over the state-of-the-arts, in line with our purpose of optimizing the margin to generate the sample with adaptive adaptability to Q (Sec.\ref{refining_eq5}). Impressively, AdaDFQ achieves at most 10.46\%, 12.59\% and 36.93\% accuracy gains on CIFAR-10, CIFAR-100 and ImageNet. Notably, compared with GDFQ, ARC and Qimera where Q is independent of the generation process, AdaDFQ obtains accuracy improvement with a large margin, \emph{e.g.}, at least 0.35\% gain (ResNet-20 with 5w5a on CIFAR-10), confirming the necessity of AdaDFQ over the sample adaptability to Q in Sec.\ref{intro}.   Specifically, without regard for the sample adaptability, ZAQ suffers from a large performance loss caused by many unexpected generated samples, which are harmful to the calibration process of Q.  AdaDFQ upgrades beyond AIT despite of the combination with ARC. As expected, AdaDFQ exhibits the obvious advantages over AdaSG, implying the benefits of optimizing the margin upon the zero-sum game.
2) AdaDFQ delivers the substantial gains for Q under varied bit-widths, confirming the importance of adaptive adaptability to varied Q (Sec.\ref{refining_eq5}). Especially for 3-bit situation, most of the competitors suffer from the poor accuracy or convergence, while AdaDFQ obtains at most 36.93\% (ResNet-18 with 3w3a) accuracy gains.

\begin{figure}[t]
\setlength{\abovecaptionskip}{0.01cm}
\setlength{\belowcaptionskip}{-0.25cm}
\centering
\includegraphics[width=0.85\columnwidth]{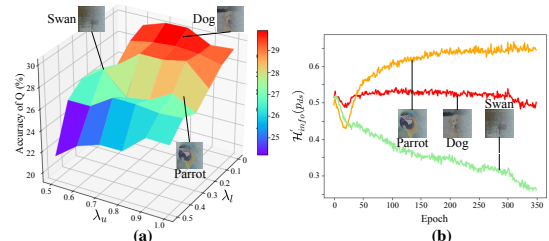}
\caption{Ablation study about $\lambda_l$ and $\lambda_u$. (a) Accuracy (\%) comparison of Q with varied ($\lambda_l$, $\lambda_u$). (b) $\mathcal{H}^{'}_{info}(p_{ds})$ for the generated samples corresponding to different areas in (a). }
\label{lu_3dmap}
\end{figure}

\begin{figure*}[t]
\centering
\includegraphics[width=1.0\textwidth]{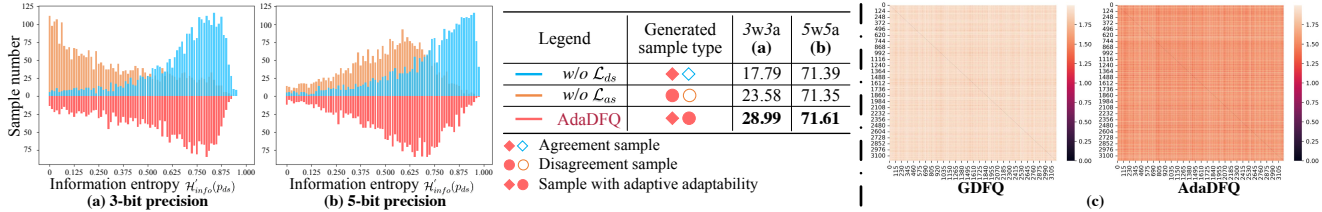}
\caption{(a)(b) Illustration of how to balance $\mathcal{L}_{ds}$ and $\mathcal{L}_{as}$ to generate the sample with adaptive adaptability under 3-bit and 5-bit precision. (c) Visual analysis: the similarity comparison between the generated samples.}
\label{experiment_sample_dist}
\end{figure*}

\subsection{Ablation Study}

\subsubsection{Validating adaptability with disagreement and agreement samples}
As aforementioned, the disagreement and agreement samples play a critical role in addressing the over-and-under fitting issues for the adaptive adaptability. We perform the ablation study on $\mathcal{L}_{ds}$ (Eq.(\ref{ds_ce})) and $\mathcal{L}_{as} $ (Eq.(\ref{las})) over ImageNet. Table \ref{ablation} suggests the noticeable superiority (38.10\%, 70.29\%) of AdaDFQ (including the both) over other cases. Note that, abandoning either or both of $\mathcal{L}_{ds}$ and $\mathcal{L}_{as} $ receives a large accuracy loss (at most 19.57\% and 4.02\%), implying the intuition of balancing the disagreement sample with agreement sample (Sec.\ref{balance_ds_as}).
Interestingly, the case without $\lambda_l$ and $\lambda_u$ (Eq.(\ref{rewritten_eq10})) obtains the minimal accuracy loss (5.97\% and 0.23\%), confirming the importance of the bound constraints on the basis of $\mathcal{L}_{bal}$ (Eq.(\ref{L_reg})).

\subsubsection{Why can $\lambda_l$ and $\lambda_u$ benefit Q?}
The parameters $\lambda_l$ and $\lambda_u$ in Eq.(\ref{rewritten_eq10}) serve as a lower and upper bound of the adaptive adaptability (neither too small or large $\mathcal{H}^{'}_{info}(p_{ds})$) for the sample generation, which is critical to address the over-and-under fitting issues.
We aim to verify the effectiveness of varied parameter configurations $(\lambda_l\in \{0,0.1,0.2,0.3,0.4,0.5\}, \lambda_u\in \{0.5,0.6,0.7,0.8,0.9,1.0\}$) via the grid search and perform the experiments under 3-bit precision with MobileNetV2 serving as P and Q on ImageNet.
Fig.\ref{lu_3dmap}(a) illustrates that AdaDFQ achieves the significant performance within an optimal range (the red area in Fig.\ref{lu_3dmap}(a)), \emph{i.e.}, $\lambda_l \in\{0,0.1,0.2\}$ and $\lambda_u \in\{0.7,0.8,0.9\}$, indicating a wide range between two bounds, where the performance of Q is insensitive to $\lambda_l$ and $\lambda_u$, which offers a guideline for the selection of their values ($\lambda_l$ and $\lambda_u$ is set to 0.1 and 0.8 in the main experiments), verifying the feasibility to uniformly select the values of $\lambda_l$ and $\lambda_u$ for all samples (Sec.\ref{lu_boundary}).
Besides, Fig.\ref{lu_3dmap}(b) provides evidence that $\lambda_l$ and $\lambda_u$ within the optimal range contributes to yielding the adaptive adaptability, where $\mathcal{H}^{'}_{info}(p_{ds})$ (the red in Fig.\ref{lu_3dmap}(b)) is neither too small (the green in Fig.\ref{lu_3dmap}(b)) nor large (the orange in Fig.\ref{lu_3dmap}(b)).

\subsubsection{How to balance disagreement sample with agreement sample?}
\label{experiment_cooper}
We further study the effectiveness of $\mathcal{L}_{bal}$ in Eq.(\ref{L_reg}), and how to balance disagreement sample with agreement sample via two cases: \textbf{A}: only generating disagreement sample, denoted as  \emph{w/o} $\mathcal{L}_{as}$; and \textbf{B}: only generating agreement sample, denoted as \emph{w/o} $\mathcal{L}_{ds}$. We perform the experiments and generate 3200 samples under 3-bit and 5-bit precision with MobileNetV2 serving as P and Q on ImageNet.
Fig.\ref{experiment_sample_dist}(a)(b) illustrates that most of the generated samples (\textcolor[RGB]{239,138,71}{---} in Fig.\ref{experiment_sample_dist}(a)(b)) from case A yield smaller $\mathcal{H}_{info}^{'}(p_{ds})$ than those (\textcolor[RGB]{255,0,0}{---} in Fig.\ref{experiment_sample_dist}(a)(b)) from case B, which provides a basis for balancing $\mathcal{L}_{ds}$ with $\mathcal{L}_{as}$. It is observed that $\mathcal{H}_{info}^{'}(p_{ds})$ of the generated sample (\textcolor[RGB]{0,176,240}{---} in Fig.\ref{experiment_sample_dist}(a)(b)) from AdaDFQ is neither too small nor large compared to case \textbf{A} and \textbf{B}, which is evidence that $\mathcal{L}_{bal}$ forces the disagreement and agreement samples to move towards each other between two boundaries, in line with the analysis in Sec.\ref{balance_ds_as}.

\begin{figure}[t]
\centering
\setlength{\abovecaptionskip}{0.02cm}
\setlength{\belowcaptionskip}{-0.35cm}
\includegraphics[width=1.0\columnwidth]{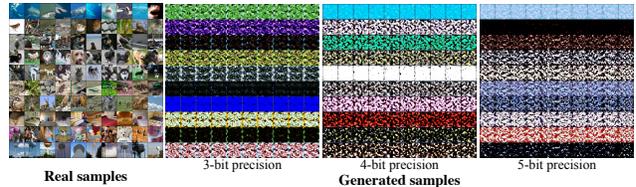}
\caption{Visualization of real and generated samples, where each row denotes one of 10 randomly chosen classes from ImageNet.}
\label{simi-visual}
\end{figure}

\subsection{Visual Analysis on Generated Samples}
To further show the intuition of the generated samples with adaptive adaptability to Q, we perform the visual analysis over MobileNetV2 serving as both P and Q on ImageNet by the similarity matrix (calculated as the $\ell_1$ norm between $p_{ds}$ of generated samples), along with the examples of generated samples (10 images per category) from 10 categories. Fig.\ref{experiment_sample_dist}(c) illustrates that the generated samples by AdaDFQ exhibit a much larger similarity (the darker, the larger) than those by GDFQ \cite{xu2020generative}, implying that the substantial samples with undesirbale adaptability by GDFQ exist against AdaDFQ. Fig.\ref{simi-visual} shows that the generated samples for different bit widths (\emph{i.e.}, 3 bit, 4 bit and 5bit) vary greatly, confirming the intuition of AdaDFQ --- generating the sample with adaptive adaptability to varied Q (Sec.\ref{refining_eq5}); while the samples from varied categories differ greatly from each other, confirming that the category information is fully exploited (Sec.\ref{balance_ds_as}); \emph{due to page limitation, see supplementary material for higher resolution}.

\section{Conclusion}
In this paper, we propose an Adaptive Data-Free Quantization (AdaDFQ) method, which revisits the DFQ from a zero-sum game perspective between two players. Following this viewpoint, the disagreement and agreement samples are further defined to form the lower and upper boundaries. The margin between two boundaries is optimized to address the over-and-under fitting issues, so as to generate the samples with the adaptive adaptability between these two boundaries to calibrate Q.
The theoretical analysis and empirical studies validate the advantages of AdaDFQ to the existing arts.

\section{Acknowledgments}
This work is supported by National Natural Science Foundation of China under the grant no U21A20470, 62172136, 72188101, U1936217. Key Research and Technology Development Projects of Anhui Province (no.202004a5020043).

{\small
\bibliographystyle{ieee_fullname}
\balance
\bibliography{egbib}
}

\end{document}